\pdfoutput=1

\documentclass[11pt]{article}

\PassOptionsToPackage{table}{xcolor}
\usepackage{ACL2023}

\usepackage[utf8]{inputenc}

\usepackage{graphicx}
\usepackage{times}
\usepackage{latexsym}
\usepackage{comment}
\usepackage{csquotes}
\usepackage{booktabs}
\usepackage{caption}
\usepackage{multirow}
\usepackage{array}
\usepackage{pifont}
\usepackage{soul}
\usepackage[normalem]{ulem}
\usepackage{makecell}
\usepackage{cprotect}
\usepackage{float}
\usepackage{subfigure}
\usepackage{hyperref}
\usepackage{listings}
\usepackage{amssymb}
\usepackage{xparse}

\NewDocumentCommand{\codeword}{v}{%
\texttt{\textcolor{blue}{#1}}%
}
\usepackage[T1]{fontenc}

\usepackage{microtype}

\usepackage{inconsolata}

\title{A Systematic Review on the Evaluation of Large Language Models in Theory of Mind Tasks}

\author{Karahan Sarıtaş \quad Kıvanç Tezören   \quad Yavuz Durmazkeser\\
\\
University of Tübingen \\
karahan.saritas@student.uni-tuebingen.de \\ kivanc.tezoeren@student.uni-tuebingen.de  \\ yavuz.durmazkeser@student.uni-tuebingen.de }

\begin{document}
\maketitle

\begin{abstract}
In recent years, evaluating the Theory of Mind (ToM) capabilities of large language models (LLMs) has received significant attention within the research community. As the field rapidly evolves, navigating the diverse approaches and methodologies has become increasingly complex. This systematic review synthesizes current efforts to assess LLMs' ability to perform ToM tasks—an essential aspect of human cognition involving the attribution of mental states to oneself and others. Despite notable advancements, the proficiency of LLMs in ToM remains a contentious issue. By categorizing benchmarks and tasks through a taxonomy rooted in cognitive science, this review critically examines evaluation techniques, prompting strategies, and the inherent limitations of LLMs in replicating human-like mental state reasoning. A recurring theme in the literature reveals that while LLMs demonstrate emerging competence in ToM tasks, significant gaps persist in their emulation of human cognitive abilities.
\end{abstract}

\section{Introduction}
Theory of Mind (ToM), the term coined by \citet{Premack1978} in their seminal paper ``\textit{Does the chimpanzee have a theory of mind?}'', refers to the cognitive ability to attribute mental states—beliefs, intents, desires, emotions, and knowledge—to oneself and others and to understand that others have mental states that are different from one's own. This is a crucial aspect of human cognition that underpins complex social interactions and empathy \cite{hofmann2016training}. By comprehending the mental states of others, an individual can infer the reasons behind observable behaviors and anticipate future actions \cite{VONK202251}. Theory of Mind has been one of the most studied fields in the developmental cognitive science \cite{Beaudoin2020}. Inspired by the Theory of Mind, researchers have developed various approaches aimed at creating models that simulate human-like learning and reasoning processes \cite{rabinowitz2018machinetheorymind, lake2016buildingmachineslearnthink}. Recent advancements in large language models (LLMs) have further intensified interest in exploring their capabilities for Theory of Mind tasks. These modern models, with their sophisticated language processing abilities, are being investigated for their potential to replicate aspects of human mental state understanding \cite{ullman2023largelanguagemodelsfail, kosinski2024evaluatinglargelanguagemodels, van-duijn-etal-2023-theory, ma-etal-2023-towards-holistic}. 
 
Large Language Models, such as GPT-3 and GPT-4, have demonstrated remarkable proficiency in understanding and generating human-like text \cite{openai2024gpt4technicalreport, brown2020languagemodelsfewshotlearners}. These models are trained on vast amounts of data and use deep learning techniques to predict and generate text based on input. Several works have been conducted to evaluate the capabilities of modern language models from a cognitive science perspective \cite{mahowald2024dissociatinglanguagethoughtlarge, dasgupta2024languagemodelshumanlikecontent}. Although large language models have achieved tremendous success at a variety of tasks such as text summarization \cite{zhang2023benchmarkinglargelanguagemodels}, question answering \cite{lewis2021retrievalaugmentedgenerationknowledgeintensivenlp}, and multilingual machine translation \cite{zhu-etal-2024-multilingual}, their performance on Theory of Mind tasks presents a new challenge. These models have demonstrated proficiency in generating coherent text, but evaluating their ability to simulate or comprehend mental states involves more nuanced assessments. 

Recently, various text-based and non-linguistic Theory of Mind tests and benchmarks have been developed to assess the cognitive capabilities of large language models \cite{kosinski2024evaluatinglargelanguagemodels, ullman2023largelanguagemodelsfail, pmlr-v162-sclar22a}. Despite the existing results, much of the work has been met with skepticism, with subsequent studies often challenging the conclusions of earlier research \cite{ullman2023largelanguagemodelsfail}. Some attribute the so-called successes of LMs in ToM tasks to an imitation of the cognitive abilities rather than really \textit{possessing} them, a long-standing debate tracing back to \citet{10.1093/mind/LIX.236.433}. Others suggest that models capable of solving these tasks often rely on memorization and shallow heuristics \cite{shapira2023cleverhansneuraltheory}, a phenomenon sometimes referred to as the \enquote{Clever Hans} effect \cite{pfungst1911clever} or the \enquote{Stochastic Parrot} problem \cite{bender2021dangers}.

\citet{kosinski2024evaluatinglargelanguagemodels} presents a dilemma in ToM evaluation of the large language models, concisely summarized by \citet{ullman2023largelanguagemodelsfail}: We must either accept the validity of current ToM measures, which would imply that large language models possess Theory of Mind, or reject the assertion that LLMs understand others' mental states, necessitating a comprehensive reevaluation and possibly a dismissal of these measures. \citet{ullman2023largelanguagemodelsfail} suggests an alternative perspective, claiming that while we can acknowledge the validity of the ToM measures, we should remain skeptical of a model that passes them. As the debate intensifies, our primary motivation for this study is to provide a systematic review of the evaluation methods used to assess the Theory of Mind in large language models. By providing a comprehensive review, we hope to clarify the current state of research, highlight areas needing further investigation, and contribute to a more nuanced understanding of how LLMs relate to human-like mental state reasoning. This systematic review will be a valuable resource for researchers and practitioners seeking to navigate the complexities of evaluating the Theory of Mind in artificial intelligence. The contributions in this paper are as follows:
\vspace{-0.5em}
\begin{itemize}
\itemsep-0.3em
\item We expand upon the existing table of benchmarks and task formulations for Large Language Models in Theory of Mind by building on the foundation provided by \citet{cao-2021-holistic}, categorizing tasks based on the taxonomy outlined by \citet{Beaudoin2020} and the concept of situatedness as highlighted by \citet{cao-2021-holistic}.

\item We distill the key findings from the papers analyzed in our review and organize them according to their primary focus areas. Drawing on the statistical outcomes, we highlight some of the major issues present in the literature.

\item We offer an in-depth overview of widely used evaluation metrics, prompting techniques, systematic failures uncovered through error analysis, and the key challenges in evaluating LLMs, all derived from our comprehensive review.
\end{itemize}

The rest of the paper is organized as follows. In Section \ref{taxonomy}, we discuss the Theory of Mind categories and sub-abilities outlined by \citet{Beaudoin2020}. Section \ref{evaluation} is divided into two parts: first, we extend the benchmark/task formulation table provided by \citet{cao-2021-holistic}, incorporating new datasets and a more detailed task categorization with links to shared resources in \ref{benchmarks}; then, we provide a brief overview of the evaluation metrics commonly used in Machine Theory of Mind in \ref{metrics}. In Section \ref{review}, we review a comprehensive list of papers in the literature, categorizing them by their focus and summarizing key findings. Section \ref{discussion} covers common prompting techniques in \ref{prompt}, the use of fine-tuning in LLMs for ToM in \ref{finetuning}, common failure reasons identified in error analysis in \ref{failure}, and major evaluation challenges in \ref{challenges}. Finally, Section \ref{conclusion} concludes the paper.

\section{Theory of Mind Taxonomy}
\label{taxonomy}

Children progressively develop a Theory of Mind through the acquisition of various cognitive components from infancy to early adolescence \cite{FU2023101061}. Initial forms of ToM start emerging in the first year of life, a period referred to as the 9-month revolution \cite{tomasello1999cultural}. They develop perception-goal psychology \cite{wellman2011childhood}, which is the ability to track others' perceptions of their surroundings and understand how they pursue their goals in response to environmental factors \cite{rakoczy2022foundations}. Around the age of 4, children experience a significant developmental shift known as the 4-year revolution, where they develop a full-fledged theory of mind. This newfound ability allows them to understand that others can hold beliefs that differ from reality, enabling them to succeed in tasks like false-belief tests. For instance, they can now recognize that someone might search for an object where they falsely believe it to be, even if the child knows its true location \cite{rakoczy2022foundations}.

Consequently, most Theory of Mind experiments have evaluated children's ability to understand different categories of mental states and social interactions. As a result of a comprehensive meta-analysis of ToM studies focused on children aged 0-5, \citet{Beaudoin2020} introduced a new taxonomy of ToM sub-domains known as Abilities in Theory of Mind Space (ATOMS). The Theory of Mind Space consists of seven primary TOM categories: intentions, desires, emotions, knowledge, percepts, beliefs, and understanding of non-literal communication. 

The \textbf{Emotions} category in the context of Theory of Mind refers to the ability to understand, infer, and explain the emotional reactions and experiences of others. This includes recognizing typical \cite{Knafo2009Empathy} and atypical \cite{Denham1986Social} emotional responses to situations, understanding that people may have mixed or hidden emotions \cite{Gordis1989Young}, and grasping the moral and regulatory aspects of emotions. The category encompasses both specific sub-abilities, such as affective perspective-taking and emotion regulation \cite{Pons2000Test}, and broader measures that assess emotional understanding across various factors like desires, beliefs, and hidden emotions \cite{Beaudoin2020}.

The \textbf{Desires} category constitutes a critical category within ATOMS, focusing on the understanding of how desires influence emotions, actions, and decision-making processes. It includes recognizing that people can have conflicting desires \cite{Repacholi1997Early}, understanding that an individual can hold multiple desires at once \cite{Bennett1993Children}, grasping how desires shape emotions and actions \cite{Wellman1988Young}, and explaining situations where actions contradict stated desires \cite{Colonnesi2008Precursors}. These aspects collectively help children navigate social interactions by understanding the role of desires in behavior.

The \textbf{Intentions} category focuses on understanding and interpreting the intentions behind others' actions. It includes recognizing intentions by completing failed actions (Behavioral Re-enactment Procedure) \cite{Meltzoff1995Understanding}, distinguishing that the same actions can result from different intentions (Accidental Transgression Task) \cite{Killen2011Accidental}, predicting actions based on intentions \cite{Phillips2002Infants}, attributing intentions to ambiguous visual figures (Valley Task) \cite{Castelli2006Valley}, and providing plausible explanations for social events based on perceived intentions \cite{Smiley2001Intention}. These aspects collectively enable individuals to interpret and anticipate the behavior of others based on their underlying motives.

The \textbf{Percepts} category in Theory of Mind involves understanding how different perceptual experiences affect individuals' actions and perspectives. This includes simple visual perspective taking, where one adopts another person's visual viewpoint as measured by tasks like the Picture Identification Task \cite{Masangkay1974Early}. It also encompasses complex visual perspective-taking, which involves more intricate tasks requiring mental rotation or visualization \cite{Ebersbach2011Relationship}. Additionally, it involves understanding the link between perception and action, where actions are connected to visual perceptions \cite{Hadwin1997Teaching}, and auditory perspective taking, which involves considering another person's auditory experiences \cite{Williamson2015Sound}. These aspects collectively help in grasping how perceptual differences influence understanding and interaction.

The \textbf{Knowledge} category involves comprehending how knowledge intersects with various cognitive and social processes. It focuses on understanding the absence of knowledge in ``pretend-play'' (Sarah task) \cite{Aronson1999Preschoolers}, understanding the effect of perceptual information on obtaining knowledge \cite{Ruffman1989Children}, having awareness about the knowledge of the reader \cite{Peskin2014Keeping}, and noting that new information is generally more captivating than what is already known \cite{Moll2006Infants}. In summary, the Knowledge category underscores the importance of recognizing how knowledge influences and is influenced by cognitive processes and social interactions, from pretend play and perceptual access to the engagement with new information.

The \textbf{Beliefs} category involves understanding and predicting the beliefs held by others, particularly when those beliefs are incorrect or based on incomplete information. It includes recognizing content false beliefs where someone is misled by the content of a container they have not inspected \cite{Hogrefe1986Ignorance}, location false beliefs where an individual does not know about a location change \cite{Wimmer1983Beliefs}, and identity false beliefs where something’s appearance leads to mistaken identity beliefs \cite{Flavell1986Development}. It also encompasses second-order beliefs, where individuals understand others' beliefs about third parties' knowledge \cite{Perner1985John}, predicting people's beliefs based on their actions \cite{Swettenham1996Children}, and sequence false beliefs caused by unexpected disruptions in predictable sequences \cite{Brambring2010Validity}. These aspects collectively illustrate how individuals predict beliefs in various contexts.

The \textbf{Non-literal Communication} category involves understanding various forms of communication that deviate from literal meaning. It includes recognizing irony and sarcasm, where people may lie or exaggerate to convey a different message \cite{Sullivan1995Children}, egocentric lies, where someone lies to avoid trouble or gain advantage \cite{Happe1994Advanced} and faux-pas (social gaffe) \cite{BaronCohen1999Recognition}. It also encompasses white lies, which are told to spare others' feelings, and involuntary lies, where individuals may inadvertently convey falsehoods and humor where lies are used to create jokes \cite{Happe1994Advanced}. This category highlights the complexity of understanding communication that goes beyond straightforward truth-telling.

\section{LLM Evaluation for ToM Abilities}
\label{evaluation}

\subsection{Theory of Mind Benchmarks for LLMs}
\label{benchmarks}

As the ToM capabilities of LLMs are being explored, the need for comprehensive evaluation frameworks becomes more prominent. \citet{ma-etal-2023-towards-holistic} provides a taxonomized review of ToM benchmarks by categorizing them with respect to the ATOMS framework. This review highlights significant gaps in current benchmarks, particularly their limited focus on specific mental states and the frequent reliance on textual data, which may lead to data contamination and shortcut learning.

Table~\ref{table::benchmarks} builds upon the foundational work of \citet{ma-etal-2023-towards-holistic}, expanding the scope to include additional benchmarks, evaluation methods, and task formulations that reflect the latest advancements in the field. We have adapted the list curated by \citet{ma-etal-2023-towards-holistic} by critically examining the existing benchmarks, incorporating our findings, and offering a more detailed taxonomy of the tasks described by these benchmarks. We modified parts that we deemed to be incorrectly categorized. For example, the input text for \citet{wu-etal-2024-coke} has been updated from \enquote{AI} to \enquote{H, AI} because the authors explicitly state that they recruited educated workers majoring in psychology for the manual selection and revision of the dataset. Our additions to the list of benchmarks include works we have collected via our search method described in Section~\ref{review}, as well as benchmarks provided in the GitHub repository\footnote{\label{ma-etal-github}\url{https://github.com/Mars-tin/awesome-theory-of-mind}} of the original table's authors, but not on the original table itself. 

\paragraph{Task Classification.} For the classification of tasks, we decided to use the following categories: Multiple Choice, True/False, Natural Language Generation, Question Answering, Inference, Text Completion, and Multi-agent Collaboration. \textit{Multiple Choice} refers to tasks where the possible answers are listed as options for the model. This is commonly used in benchmarks and task formulations \cite{ma2023tomchallengesprincipleguideddatasetdiverse, mireshghallah2024llmssecrettestingprivacy, kim2023fantombenchmarkstresstestingmachine, gandhi2023understandingsocialreasoninglanguage}. \textit{True/False} consists of questions where the model is expected to respond with true/yes/1 or false/no/0. This question format is also preferred due to its simplicity \cite{van-duijn-etal-2023-theory, ma2023tomchallengesprincipleguideddatasetdiverse}. \textit{Natural Language Generation} tasks expect longer, open-ended responses, where the generated output is usually evaluated by comparing it to a set of reference texts. One example of a Natural Language Generation (NLG) task in \citet{mireshghallah2024llmssecrettestingprivacy} involves prompting the model to answer a question while considering privacy norms. The detection of leakage in the response is then assessed using two methods: (a) exact string matching for X’s name, and (b) determining whether a proxy model can recover the private information solely from the given response. 
\textit{Question Answering} tasks require the model to respond to a question with the most appropriate answer. This category typically involves extracting information from a given text or reasoning about the content to generate the correct response. It differs from \textit{Multiple Choice} tasks, as it does not provide predefined options, and from \textit{Natural Language Generation} tasks, as it usually expects a concise answer, often a single word or a short phrase, rather than a longer, open-ended response. Examples of this category can be found in \citet{ma2023tomchallengesprincipleguideddatasetdiverse, xu2024opentomcomprehensivebenchmarkevaluating, chan2024negotiationtombenchmarkstresstestingmachine}. 
\textit{Text Completion} involves tasks where the model is expected to complete a missing part of a sentence, often used for base language models \cite{van-duijn-etal-2023-theory}. \textit{Multi-agent Collaboration} consists of multiplayer tasks where agents work together to solve a specific problem. These tasks may have different objective functions depending on the nature of the game \cite{bianchi2024llmsnegotiatenegotiationarenaplatform, Li_2023, bara2021mindcraft, guo2023suspicionagentplayingimperfectinformation, pmlr-v162-sclar22a} Lastly, \textit{Inference} refers to tasks where the model is expected to make logical inferences, which may involve Natural Language Inference \cite{cohen2021exploring} or predictions using methods like logistic regression \cite{eysenbach2016mistaken}.

\newcommand{\cmark}{\ding{51}}
\begin{table*}[ht]
\centering
\vspace{-8pt}
\scalebox{0.58}{
\hspace*{-2.1cm}
\begin{tabular}{lcccccccccccccccccc}
    \toprule
    \multirow{3}{*}{Benchmarks and Task Formulations} & \multirow{3}{*}{\makecell{Resources}} & \multicolumn{3}{c}{Tested Agent} &  \multicolumn{4}{c}{Situatedness} & \multicolumn{9}{c}{ATOMS Mental States} & \multirow{3}{*}{Sym.} \\
    \cmidrule(r){3-5} \cmidrule(r){6-9} \cmidrule(r){10-18} & & \multirow{2}{*}{Task} & \multicolumn{2}{c}{Input Modality} & \multicolumn{2}{c}{Physical} & \multicolumn{2}{c}{Social} & \multicolumn{2}{c}{Belief} & \multicolumn{2}{c}{Intention} & \multirow{2}{*}{Des.} & \multirow{2}{*}{Emo.} & \multirow{2}{*}{Know.} & \multirow{2}{*}{Per.} & \multirow{2}{*}{NLC} & \\
    & & & Text & Nonling. & Per. & Int. & Per. & Int. & 1st & 2nd+ & Act. & Com. & & & & & & \\
    \cmidrule(r){1-1} \cmidrule(r){2-2} \cmidrule(r){3-5} \cmidrule(r){6-9} \cmidrule(r){10-18} \cmidrule(r){19-19}
    \rowcolor[HTML]{DAE8FC}
    \textsc{\textdagger Epistemic Reasoning}~{\small\protect\citep{cohen2021exploring}} & - & Infer & T & \multicolumn{1}{c|}{-} & & & & \multicolumn{1}{c|}{} & \cmark & \cmark & & & & & & & & \multicolumn{1}{|c}{} \\
    \textsc{ToMi}~{\small\protect\citep{nematzadeh2018evaluating}} & \protect\href{https://github.com/facebookresearch/ToMi}{Code/Data} & QA & T & \multicolumn{1}{c|}{-} & \cmark & & & \multicolumn{1}{c|}{} & \cmark & \cmark & & & & & & & & \multicolumn{1}{|c}{} \\
    \rowcolor[HTML]{DAE8FC}
    \textsc{Hi-ToM}~{\small\protect\citep{wu-etal-2023-hi}} & \protect\href{https://github.com/ying-hui-he/hi-tom_dataset}{Code/Data} & MC & T & \multicolumn{1}{c|}{-} & \cmark & & & \multicolumn{1}{c|}{} & \cmark & \cmark & & & & & & & & \multicolumn{1}{|c}{} \\
    \textsc{MindGames}~{\small\protect\citep{sileo-lernould-2023-mindgames}} & \protect\href{https://github.com/sileod/llm-theory-of-mind}{Code}, \protect\href{https://huggingface.co/datasets/sileod/mindgames}{Data} & Infer, T/F & T & \multicolumn{1}{c|}{-} & \cmark & & & \multicolumn{1}{c|}{} & \cmark & \cmark & & & & & & \cmark & & \multicolumn{1}{|c}{} \\
    \rowcolor[HTML]{DAE8FC}
    \textsc{Adv-CSFB}~{\small\protect\citep{shapira2023cleverhansneuraltheory}} & \protect\href{https://github.com/salavi/Clever_Hans_or_N-ToM}{Data} & MC, Infer, TC & H & \multicolumn{1}{c|}{-} & \cmark & & & \multicolumn{1}{c|}{} & \cmark & & & & & & & & & \multicolumn{1}{|c}{} \\
    \textsc{\textdagger ConvEntail}~{\small\protect\citep{zhang2010towards}} & \protect\href{https://github.com/lairmsu/ConversationalEntailment}{Data} & Infer & H & \multicolumn{1}{c|}{-} & & & \cmark & \multicolumn{1}{c|}{} & \cmark & & & \cmark & \cmark & & & & & \multicolumn{1}{|c}{} \\
    \rowcolor[HTML]{DAE8FC}
    \textsc{SocialIQA}~{\small\protect\citep{sap-etal-2019-social}} & \protect\href{https://maartensap.com/social-iqa/}{Data} & MC & H & \multicolumn{1}{c|}{-} & & & \cmark & \multicolumn{1}{c|}{} & & & \cmark & & & \cmark & & & & \multicolumn{1}{|c}{} \\
    \textsc{BeSt}~{\small\protect\citep{tracey2022best}} & - & - & H & \multicolumn{1}{c|}{-} & & & \cmark & \multicolumn{1}{c|}{} & \cmark & & & & & \cmark & & & \cmark & \multicolumn{1}{|c}{} \\
    \rowcolor[HTML]{DAE8FC}
    \textsc{FauxPas-EAI}~{\small\protect\citep{shapira2023how}} & \protect\href{https://github.com/NatalieShapira/FauxPasEAI}{Code/Data} & NLG & H,AI & \multicolumn{1}{c|}{-} & & & \cmark & \multicolumn{1}{c|}{} & \cmark & & & & & & & & \cmark & \multicolumn{1}{|c}{} \\
    \textsc{COKE}~{\small\protect\citep{wu-etal-2024-coke}} & \protect\href{https://github.com/jincenziwu/COKE}{Code/Data} & NLG & H, AI & \multicolumn{1}{c|}{-} & & & \cmark & \multicolumn{1}{c|}{\cmark} & & & \cmark & & & \cmark & & & & \multicolumn{1}{|c}{} \\
    \rowcolor[HTML]{DAE8FC}
    \textsc{ToM-in-AMC}~{\small\protect\citep{yu2024fewshotcharacterunderstandingmovies}} & \protect\href{https://github.com/Gorov/tom_in_amc}{Data}& MC, NLG & H & \multicolumn{1}{c|}{-} & \cmark & & \cmark & \multicolumn{1}{c|}{} & & & \cmark & \cmark & & & & & & \multicolumn{1}{|c}{} \\
    \textsc{G4C}~{\small\protect\citep{zhou2023i}} & - & NLG & H,AI & \multicolumn{1}{c|}{-} & \cmark & & \cmark & \multicolumn{1}{c|}{\cmark} & & & \cmark & \cmark & & & & \cmark & & \multicolumn{1}{|c}{} \\
    \rowcolor[HTML]{DAE8FC}
    \textsc{VisualBeliefs}~{\small\protect\citep{eysenbach2016mistaken}} & \protect\href{http://people.csail.mit.edu/bce/mistaken/}{Code/Data} & Infer & - & \multicolumn{1}{c|}{Cartoon} & \cmark & & & \multicolumn{1}{c|}{} & \cmark & & & & & & & & \cmark & \multicolumn{1}{|c}{} \\
    \textsc{Triangle COPA}~{\small\protect\citep{gordon2016commonsense}} & \protect\href{https://github.com/asgordon/TriangleCOPA}{Data} & QA & H & \multicolumn{1}{c|}{Cartoon} & \cmark & & \cmark & \multicolumn{1}{c|}{} & & & \cmark & & & \cmark & & & & \multicolumn{1}{|c}{} \\
    \rowcolor[HTML]{DAE8FC}
    \textsc{MSED}~{\small\protect\citep{jia2022beyond}} & \protect\href{https://github.com/MSEDdataset/MSED}{Data} & Infer & H & \multicolumn{1}{c|}{Images} & \cmark & & & \multicolumn{1}{c|}{} & & & & & \cmark & \cmark & & & & \multicolumn{1}{|c}{} \\
    \textsc{BIB}~{\small\protect\citep{gandhi2021baby}} & \protect\href{https://www.kanishkgandhi.com/bib}{Code/Data} & Infer & - & \multicolumn{1}{c|}{2D Grid} & \cmark & & & \multicolumn{1}{c|}{} & & & \cmark & & \cmark & & & & & \multicolumn{1}{|c}{} \\
    \rowcolor[HTML]{DAE8FC}
    \textsc{AGENT}~{\small\protect\citep{shu2021agent}} & \protect\href{https://www.tshu.io/AGENT/}{Code/Data} & Infer & - & \multicolumn{1}{c|}{3D Sim.} & \cmark & & & \multicolumn{1}{c|}{} & & & \cmark & & \cmark & & & \cmark & & \multicolumn{1}{|c}{} \\
    \textsc{MToM}~{\small\protect\citep{rabinowitz2018machinetheorymind}} & - & Infer & - & \multicolumn{1}{c|}{2D Grid} & \cmark & & & \multicolumn{1}{c|}{} & \cmark & & \cmark & & & & & & & \multicolumn{1}{|c}{} \\
    \rowcolor[HTML]{DAE8FC}
    \textsc{SymmToM}~{\small\protect\citep{pmlr-v162-sclar22a}} & \protect\href{https://github.com/msclar/symmtom}{Code} & MACol & - & \multicolumn{1}{c|}{2D Grid} & \cmark & \cmark & \cmark & \multicolumn{1}{c|}{\cmark} & & & & & & & \cmark & & & \multicolumn{1}{|c}{\cmark} \\
    \textsc{MindCraft}~{\small\protect\citep{bara2021mindcraft}} & \protect\href{https://github.com/sled-group/MindCraft}{Code} & MACol & H & \multicolumn{1}{c|}{3D Sim.} & \cmark & \cmark & \cmark & \multicolumn{1}{c|}{\cmark} & & & \cmark & & & & \cmark & \cmark & & \multicolumn{1}{|c}{\cmark} \\
    \rowcolor[HTML]{DAE8FC}
    \textsc{CPA}~{\small\protect\citep{bara2023towards}} & \protect\href{https://github.com/sled-group/collab-plan-acquisition}{Code/Data} & Infer & H & \multicolumn{1}{c|}{3D Sim.} & \cmark & \cmark & \cmark & \multicolumn{1}{c|}{\cmark} & & & \cmark & \cmark & & & \cmark & \cmark & & \multicolumn{1}{|c}{\cmark} \\
    \textsc{* ToMChallenges}~{\small\protect\citep{ma2023tomchallengesprincipleguideddatasetdiverse}} & \protect\href{https://github.com/xiaomeng-ma/ToMChallenges}{Code/Data} & T/F, TC, QA, & T & \multicolumn{1}{c|}{-} & \cmark & & & \multicolumn{1}{c|}{}
        & \cmark & \cmark & & & & & & & & \multicolumn{1}{|c}{} \\
    \rowcolor[HTML]{DAE8FC}
    \textsc{* ConfAIde}~{\small\protect\citep{mireshghallah2024llmssecrettestingprivacy}} & \protect\href{https://github.com/skywalker023/confAIde}{Code/Data} & MC, NLG & T, AI & \multicolumn{1}{c|}{-} & & & \cmark & \multicolumn{1}{c|}{\cmark}
        & \cmark & & & \cmark & & & \cmark & & & \multicolumn{1}{|c}{} \\
    \textsc{* FanToM}~{\small\protect\citep{kim2023fantombenchmarkstresstestingmachine}} & \protect\href{https://github.com/skywalker023/fantom}{Code/Data} & T/F, MC, QA, NLG & H & \multicolumn{1}{c|}{-} & & & \cmark & \multicolumn{1}{c|}{}
        & \cmark & \cmark & & & & & \cmark & & & \multicolumn{1}{|c}{} \\
    \rowcolor[HTML]{DAE8FC}
    \textsc{* BigToM}~{\small\protect\citep{gandhi2023understandingsocialreasoninglanguage}} & \protect\href{https://github.com/cicl-stanford/procedural-evals-tom}{Code/Data} & MC & T, AI & \multicolumn{1}{c|}{-} & \cmark & & & \multicolumn{1}{c|}{}
        & \cmark & \cmark & & & \cmark & & & \cmark & & \multicolumn{1}{|c}{} \\
    \textsc{*\textdagger Kosinski's Tasks}~{\small\protect\citep{kosinski2024evaluatinglargelanguagemodels}} & \protect\href{https://osf.io/csdhb/}{Code/Data} & TC & H & \multicolumn{1}{c|}{-} & \cmark & & & \multicolumn{1}{c|}{}
        & \cmark & & & & & & & & & \multicolumn{1}{|c}{} \\
    \rowcolor[HTML]{DAE8FC}
    \textsc{* InterIntent}~{\small\protect\citep{liu2024interintentinvestigatingsocialintelligence}} & - & MC, NLG & T & \multicolumn{1}{c|}{-} & & & \cmark & \multicolumn{1}{c|}{}
        & & & \cmark & \cmark & & & & & & \multicolumn{1}{|c}{} \\
    \textsc{* T4D}~{\small\protect\citep{zhou2023farlargelanguagemodels}} & - & MC & T & \multicolumn{1}{c|}{-} & \cmark & & \cmark & \multicolumn{1}{c|}{\cmark}
        & \cmark & \cmark & & & & & \cmark & & \cmark & \multicolumn{1}{|c}{} \\
    \rowcolor[HTML]{DAE8FC}
    \textsc{*\textdagger Search \& Rescue}~{\small\protect\citep{Li_2023}} & - & MACol & - & \multicolumn{1}{c|}{2D Grid} & \cmark & \cmark & \cmark & \multicolumn{1}{c|}{\cmark}
        & \cmark & \cmark & & & & & \cmark & \cmark & & \multicolumn{1}{|c}{\cmark} \\
    \textsc{* Bloom}~{\small\protect\citep{leer2023violationexpectationmetacognitiveprompting}} & \protect\href{https://github.com/plastic-labs/voe-paper-eval}{Code/Data} & NLG & H, AI & \multicolumn{1}{c|}{-} & & & \cmark & \multicolumn{1}{c|}{}
        & \cmark & & & \cmark & & & & & & \multicolumn{1}{|c}{} \\
    \rowcolor[HTML]{DAE8FC}
    \textsc{* MoToMQA}~{\small\protect\citep{street2024llmsachieveadulthuman}} & - & T/F & H & \multicolumn{1}{c|}{-} & & & \cmark & \multicolumn{1}{c|}{}
        & \cmark & \cmark & & \cmark & & & & \cmark & & \multicolumn{1}{|c}{} \\
    \textsc{* Open-ToM}~{\small\protect\citep{xu2024opentomcomprehensivebenchmarkevaluating}} & \protect\href{https://seacowx.github.io/projects/opentom/OpenToM.html}{Code/Data} & QA & H, AI & \multicolumn{1}{c|}{-} & \cmark & & \cmark & \multicolumn{1}{c|}{}
        & \cmark & \cmark & \cmark & & \cmark & \cmark & & \cmark & & \multicolumn{1}{|c}{} \\
    \rowcolor[HTML]{DAE8FC}
    \textsc{* NegotiationToM}~{\small\protect\citep{chan2024negotiationtombenchmarkstresstestingmachine}} & \protect\href{https://github.com/HKUST-KnowComp/NegotiationToM}{Data} & MC, QA & H & \multicolumn{1}{c|}{-} & & & \cmark & \multicolumn{1}{c|}{}
        & & \cmark & \cmark & \cmark & \cmark & & & & & \multicolumn{1}{|c}{} \\
    \textsc{* NegotiationArena}~{\small\protect\citep{bianchi2024llmsnegotiatenegotiationarenaplatform}} & \protect\href{https://github.com/vinid/NegotiationArena}{Code} & MACol & AI & \multicolumn{1}{c|}{-} & & & & \multicolumn{1}{c|}{\cmark}
        & \cmark & \cmark & \cmark & \cmark & \cmark & \cmark & & & & \multicolumn{1}{|c}{\cmark} \\
    \rowcolor[HTML]{DAE8FC}
    \textsc{* Reddit CMV}~{\small\protect\citep{amirizaniani2024llmsexhibithumanlikereasoning}} & - & TC & H & \multicolumn{1}{c|}{-} & & & & \multicolumn{1}{c|}{\cmark}
        & & & & \cmark & & \cmark & & & & \multicolumn{1}{|c}{} \\
    \textsc{* Percept-ToMi}~{\small\protect\citep{jung2024perceptionsbeliefsexploringprecursory}} & - & QA & H & \multicolumn{1}{c|}{-} & & & \cmark & \multicolumn{1}{c|}{}
        & \cmark & \cmark & & & & & & \cmark & & \multicolumn{1}{|c}{} \\
    \rowcolor[HTML]{DAE8FC}
    \textsc{* Percept-FANToM}~{\small\protect\citep{jung2024perceptionsbeliefsexploringprecursory}} & - & QA & H & \multicolumn{1}{c|}{-} & & & \cmark & \multicolumn{1}{c|}{}
        & \cmark & \cmark & & & & & & \cmark & & \multicolumn{1}{|c}{} \\
    \textsc{* ToM-PROBE}~{\small\protect\citep{Verma_2024}} & - & T/F & H & \multicolumn{1}{c|}{-} & & & & \multicolumn{1}{c|}{}
        & & \cmark & & & & & \cmark & \cmark & & \multicolumn{1}{|c}{} \\
    \rowcolor[HTML]{DAE8FC}
    \textsc{* EmoBench}~{\small\protect\citep{sabour-etal-2024-emobench}} & \protect\href{https://github.com/Sahandfer/EmoBench}{Code/Data} & MC & H & \multicolumn{1}{c|}{-} & & & \cmark & \multicolumn{1}{c|}{\cmark}
        & \cmark & & & & & \cmark & & & \cmark & \multicolumn{1}{|c}{} \\
    \textsc{* Loophole}~{\small\protect\citep{murthy-etal-2023-comparing}} & \protect\href{https://github.com/skmur/LLLMs}{Code} & NLG & H & \multicolumn{1}{c|}{-} & & & \cmark & \multicolumn{1}{c|}{}
        & & & & & & & & & \cmark & \multicolumn{1}{|c}{} \\
    \rowcolor[HTML]{DAE8FC}
    \textsc{* RBC}~{\small\protect\citep{stohr2023deciphering}} & - & MACom & - & \multicolumn{1}{c|}{2D Grid} & \cmark & & & \multicolumn{1}{c|}{}
        & \cmark & & & & & & \cmark & & & \multicolumn{1}{|c}{} \\
    \textsc{* Selective Encoding}~{\small\protect\citep{ruis2023do}} & \protect\href{https://github.com/LauraRuis/tom}{Code/Data} & QA & T & \multicolumn{1}{c|}{-} & \cmark & & & \multicolumn{1}{c|}{}
        & & & \cmark & & \cmark & & & & & \multicolumn{1}{|c}{} \\
    \bottomrule
\end{tabular}
}
\cprotect\caption{Our extended taxonomized review of ToM benchmarks, adapted from~\protect\citet{ma-etal-2023-towards-holistic}. The table is not listed in a specific order. We retain the analyzed fields presented in the original table, and additionally include the \textbf{Resources} column. Among ATOMS categories, \textbf{Beliefs} are separated as first-order (\uline{1st}) and second-order beliefs or beyond (\uline{2nd+}). \textbf{Intentions} are likewise broken into \uline{Act}ion and \uline{Com}municative intentions. \textbf{Situatedness} is presented as None, Passive \uline{Per}ceiver, and Active \uline{Int}eractor. As in the original table, \textbf{Symmetricity} implies whether tested agents are situated together and mutually interacting with other agents. We provide a more detailed categorization for \textbf{Tasks} as \uline{M}ultiple \uline{C}hoice Question Answering, \uline{T}rue/\uline{F}alse Question Answering, Sentence \uline{Q}uestion \uline{A}nswering, \uline{T}ext \uline{C}ompletion, \uline{I}nference, \uline{N}atural \uline{L}anguage \uline{G}eneration, \uline{M}ultiple \uline{A}gent \uline{Col}laboration, and \uline{M}ultiple \uline{A}gent \uline{Com}petition. \textbf{Input Modalities} presented as in the original table: \uline{H}uman, \uline{AI} or \uline{T}emplate for \textbf{Text}, and \uline{Cartoon}, Natural \uline{Image}, \uline{2D Grid} World or \uline{3D Sim}ulation for \textbf{\uline{Nonling}uistic} input. Our additions to the original table are indicated with (*). Benchmarks without official names and whose names are given by~\protect\citet{ma-etal-2023-towards-holistic} or us and are denoted with (\textdagger).}
\label{table::benchmarks}
\end{table*}

\paragraph{Situatedness.} Despite emphasizing situatedness in their work, \citet{ma-etal-2023-towards-holistic} does not provide a detailed explanation of what constitutes a physically perceiver/interactor versus a socially perceiver/interactor in their taxonomized review. Based on the classification of existing benchmarks, we have adopted the following rules for our classification: We classify an agent as a physically perceiver if it can perceive environmental content, object locations, etc. An agent is considered a physical interactor if it can interact with objects or move around. An agent is a socially perceiver if it is informed about social circumstances in the environment, such as emotions, plans, and conversations between humans. Lastly, an agent is categorized as a social interactor if it engages with humans or other agents by generating actions.

For example, \citet{cohen2021exploring} is considered to have no situatedness at all. The agent does not perceive the environment or any actions/social situations; the dataset consists solely of natural language inference queries. \citet{wu-etal-2023-hi} presents HI-TOM, in which an agent is a physical perceiver—it can see objects in a room and observe their location changes. However, the agent does not interact with the objects, and no social actions are present in the setting (e.g., object manipulation or agent movements are not considered complex actions). \citet{sap-etal-2022-neural} introduces SocialIQA, where an agent is presented with social circumstances and is expected to reason about underlying motivations, predict future events, and understand emotional reactions. However, it does not produce any social actions to be classified as a social interactor. \citet{wu-etal-2024-coke} presents COKE, in which the agent is both a socially interactive and perceiving entity; it perceives social circumstances and generates actions and emotions. Finally, \citet{pmlr-v162-sclar22a} describes a dataset where the agent is placed in a grid, can move around, observe other agents and their actions, and interact with them socially. This dataset is also considered symmetric, as other agents can take actions as well.

Finally, we introduced a new column in our table that includes links to relevant resources, such as repositories and datasets. By expanding and updating this table, we aim to create a more robust and comprehensive resource for researchers who are investigating the Theory of Mind (ToM) capabilities of large language models (LLMs). This enhancement is crucial for addressing the complex and nuanced challenges that arise when evaluating machine Theory of Mind, especially within the context of situated and interactive environments.

\subsection{Evaluation Metrics}
\label{metrics}
Depending on the task, obtaining a quantitative value to assess the performance may pose a challenge. For benchmarks that only require giving a short answer or choosing one of the predefined options, one can compare the output with a string to determine the correctness of an example. Calculating the accuracy then becomes straightforward. One useful technique is to prompt the model to output in JSON format, making it easier to parse for automatic grading \cite{jung2024perceptionsbeliefsexploringprecursory}. Pattern matching can also be used to find specific words in the generated response \cite{kim2023fantombenchmarkstresstestingmachine}. However, we need to be careful since multiple correct answers can exist for a single question. Selecting the correct labels is important for a precise computation of the accuracy. Additionally, F1-score or Area Under the Precision-Recall Curve (AUC-PR) can be reported when there exists label imbalance \cite{kwon2024llmseffectivenegotiatorssystematic, fang2024inferactinferringsafeactions}.

For natural language generation tasks, such a direct computation of accuracy is not possible. Instead, the output is compared with a set of reference texts that are assumed to be correct. The reference sentences are written by humans, and are a part of the dataset. BLEU \cite{papineni2002bleu}, ROUGE \cite{lin2004rouge}, CIDEr \cite{vedantam2015cider}, METEOR \cite{banerjee2005meteor}, and BERTScore \cite{zhang2020bertscore} are some popular evaluation metrics. BLEU, ROUGE, CIDEr, and METEOR use n-grams for assigning a score. On the other hand, BERTScore computes the similarities between BERT \cite{devlin2019bert} embeddings of two sentences. As those metrics rely on the reference sentences, the results are affected by the quality and the diversity of the reference text. These methods also measure the alignment between the models and the human expectations.

In multi-agent competitive/collaborative games, game-specific metrics can be employed. For instance, \citet{liu2024interintentinvestigatingsocialintelligence} uses metrics such as win rate, quest win rate, quest engagement rate, team selection accuracy, failure vote rate, team proposal change rate, and Merlin assassination rate—each of which depends on the unique dynamics of the game.

Determining the appropriate metric to use or design is often challenging. To validate their automatic metrics, \citet{zhou2023i} compare them with human evaluations. By measuring the correlation between human evaluations and the results of the automatic metrics, they identify which tasks can rely on automated metrics and which require manual scoring by humans for the test set.

When a dataset contains multiple stories with multiple questions per story, \textit{joint accuracy} \cite{le-etal-2019-revisiting} can be applied. This approach requires the model to answer all questions for a given story correctly in order to earn 1 point—otherwise, it receives nothing. As the authors note, this metric ``is not inflated by easy-to-answer memory and reality questions''.

\section{Literature Review}
\label{review}

\begin{figure*}[h]
    \centering
    \subfigure{
        \includegraphics[width=0.45\textwidth]{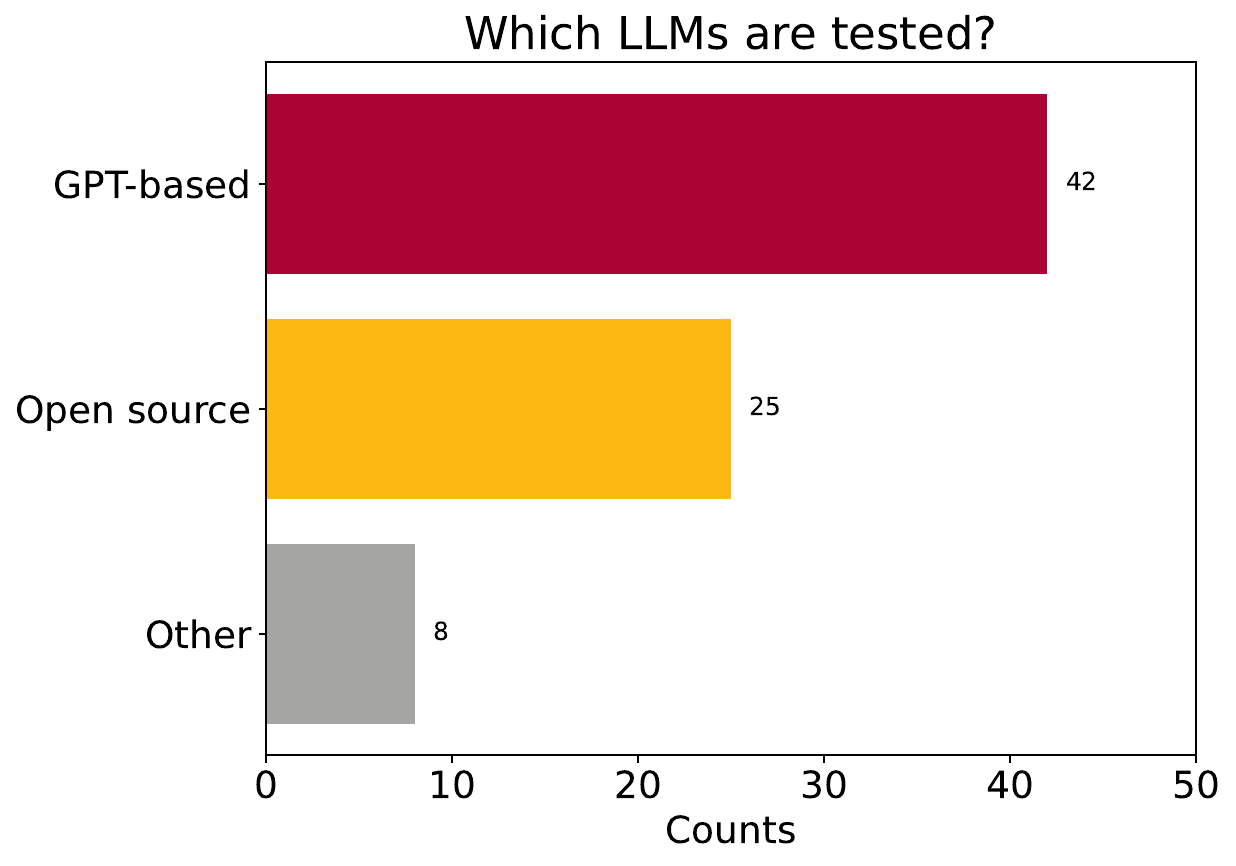}
    }
    \hspace{0.05\textwidth}
    \subfigure{
        \includegraphics[width=0.45\textwidth]{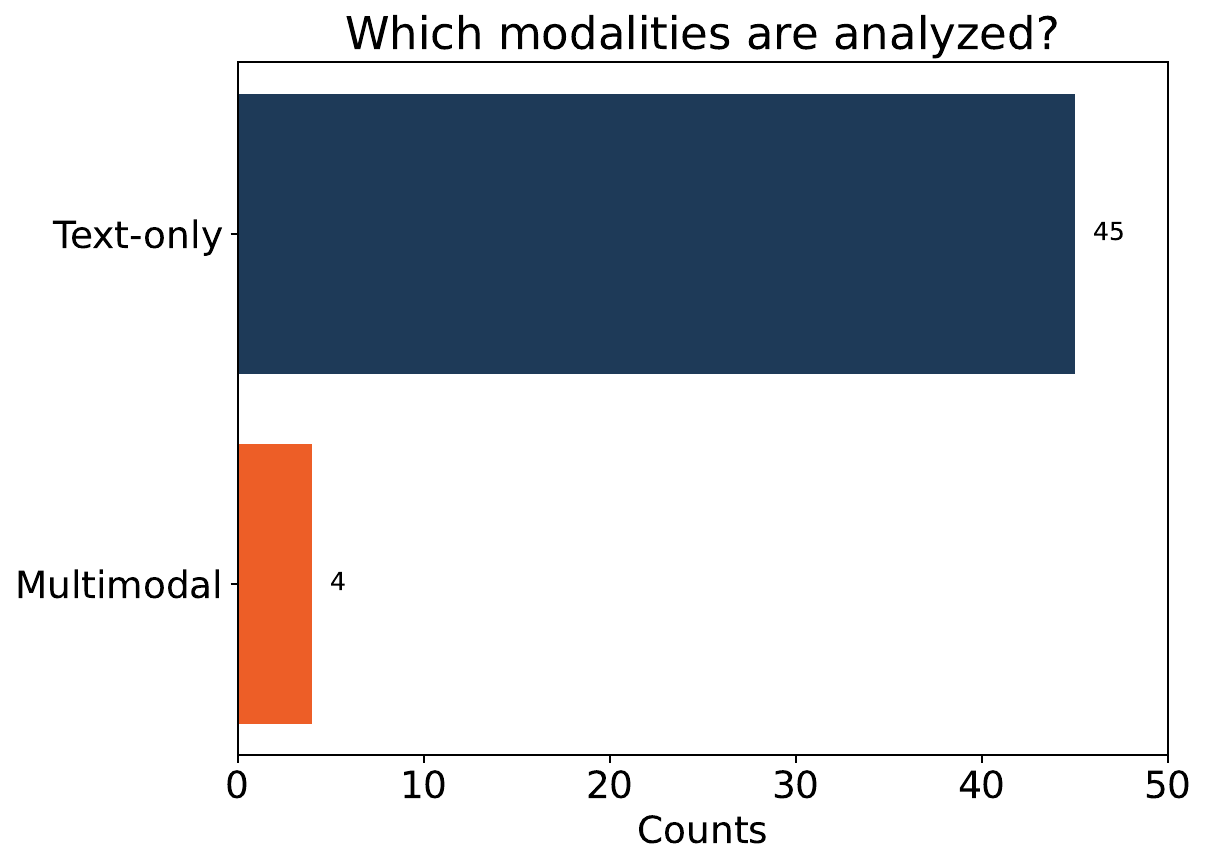}
    }
    \caption{Comparison of the number of papers inspecting different LLMs (left) and modalities (right).}
    \label{fig:literature_summary}
\end{figure*}

As a systematic ground for our review of ToM in LLMs research, we focused on ArXiv papers and searched the database using the ArXiv API\footnote{\url{https://info.arxiv.org/help/api/index.html}}. The search query was set as \codeword{"theory of mind"} \codeword{AND ("LLM" OR "large language models")} \footnote{Using "ToM" in the query retrieved unrelated papers and didn’t yield any new results compared to using only "theory of mind".}.  In addition to Arxiv API, we reviewed the Theory of Mind benchmarks listed in a repository\textsuperscript{\ref{ma-etal-github}} maintained by \citet{ma-etal-2023-towards-holistic}.
As a result, we reviewed a total of 58 papers related to the Machine Theory of Mind. Since this is a relatively recent field, we intentionally limited the number of papers to avoid including scientifically insufficient works merely to increase the quantity reviewed.
It is important to note that not all papers explicitly mention the Theory of Mind, include experiments, or use large language models in their studies—however, all of them focus on specific aspects of the Machine Theory of Mind.

Fig. \ref{fig:literature_summary} shows some insights about the investigated papers. We observed that out of 49 papers that experiment with LLMs, 42 of them test one or more GPT-based models. On the other hand, 25 of them analyzed open-source models. Finally, only 8 of them used other models with API access. Possibly due to their superior language abilities, GPT models dominate the ToM research on LLMs. Additionally, we noticed that most of the work is in the text-only domain. Only 5 out of 49 papers include other modalities than text.

We additionally investigated which LLM evaluation schemes were used by the authors when reporting performance on the desired tasks. We have found that out of the discussed 49 papers which evaluate LLMs,
\vspace{-0.5em}
\begin{itemize}
\itemsep-0.3em
    \item 39 of them (79.6\%) reported automated objective performance metrics, such as accuracy on a specified test suite with ground-truth values,
    \item 10 of them (20.4\%) reported LLM performance measures based on manual evaluation by human experts,
    \item and 17 of them (34.7\%) compared LLM performance with human performance on the same task, potentially with further statistical analysis.
\end{itemize}
Note that the listed evaluation methods are not mutually exclusive. As such, we have noticed that out of 10 papers using human evaluation, 5 of them additionally report an automated accuracy value. Ground-truth-based evaluation takes up a high portion of the analyzed papers, which highlights the importance of structured benchmarks and numerical analysis in the literature. However, it should be noted that numerical comparison with human performance is not as prevalent, and might reflect a shortcoming of the current ToM in LLMs literature.

\paragraph{False Belief Understanding.} So far the majority of the work in Machine Theory of Mind focused on the false belief tests, usually with Sally-Anne \cite{BARONCOHEN198537} and Smarties \cite{gopnik1988childrens} variants. The Smarties Test and the Sally-Anne Test are classic assessments of children's Theory of Mind. In the Smarties Test, children are shown a tube labeled as containing candy but discover it holds pencils. Younger children struggle to predict that others would still expect candy inside, highlighting their difficulty understanding that others can hold beliefs different from reality. The Sally-Anne Test involves a character who mistakenly believes her marble is where she left it, despite it being moved. Children's responses reveal whether they grasp that others can have false beliefs. Both tests also have second-order versions that further assess the child's ability to understand beliefs about beliefs. 

\citet{ma2023tomchallengesprincipleguideddatasetdiverse} develop a benchmark based on false belief tests, while \cite{van-duijn-etal-2023-theory} incorporate standardized false belief tests along with tasks involving non-literal communication (faux pas) and knowledge tests (imposing memory tests). In his much-debated work, \citet{kosinski2024evaluatinglargelanguagemodels} explores various false-belief scenarios to assess the Theory of Mind abilities of large language models, ultimately suggesting that ToM skills may have emerged as a byproduct of the training process—a claim that would later face criticism by \citet{ullman2023largelanguagemodelsfail} who showed that the small variations to false belief tasks can lead to significantly worse accuracies in tests. These variations can include uninformative additions to prompt or making the container transparent. 
In his updated work, \citet{kosinski2024evaluatinglargelanguagemodels} criticizes the methodology of \citet{ullman2023largelanguagemodelsfail}, stating, \enquote{First, claims about LLMs’ performance should be backed by empirical evidence and statistical analyses, rather than anecdotal examples. Ullman, for instance, selected two out of our 40 tasks and designed eight true-belief scenarios that GPT-3-davinci-003 failed to solve. However, a few examples of white swans (tasks a model cannot solve) neither prove that black swans (tasks a model can solve) do not exist nor do they inform us about the true white-to-black swan ratio.}

Then, \citet{shapira2023cleverhansneuraltheory} adopt a similar approach by introducing adversarial examples to the dataset, concluding that 'LLMs don’t have robust ToM abilities but rather rely on shallow heuristics.' \citet{kim2023fantombenchmarkstresstestingmachine} is among other works that focus on false-belief understanding to measure the ToM abilities of LLMs.

\paragraph{Reasoning.} \citet{amirizaniani2024llmsexhibithumanlikereasoning} compare the reasoning of LLMs with humans on Reddit CMV\footnote{\url{https://www.reddit.com/r/changemyview/}} data. They encounter some problems with ToM reasoning on open ended questions, even after prompt tuning.

\paragraph{Perception.} \citet{jung2024perceptionsbeliefsexploringprecursory} claim that LLMs can identify what a character sees or hears, but they fail to deduce what a character believes based on what they perceive. They notice that irrelevant information confuses the models, and propose a prompting method called PercepToM which guides them through a structured process of inference. \citet{kosoy2023comparingmachineschildrenusing} similarly report a case where an LLM fails in processing perception.

\paragraph{Privacy.} \citet{mireshghallah2024llmssecrettestingprivacy} poses the question, ``Can LLMs keep a secret?'' and introduces a new benchmark called ConfAIde, which consists of four tiers of increasing complexity. For example, in the final tier, the model is expected to generate a summary of a text containing private information that will be distributed to all meeting attendees. The model is required to produce a summary that addresses privacy concerns. They demonstrate that advanced models like GPT-4 and ChatGPT disclose private information in scenarios where humans would typically refrain from doing so, 39\% and 57\% of the time, respectively.

\paragraph{Social Intelligence.}
Several studies have been conducted to measure the social intelligence of language models, which necessitates a certain level of Theory of Mind. To measure social intelligence, benchmarks such as SocialIQA \cite{sap-etal-2019-social}, EmoBench \cite{sabour-etal-2024-emobench} and BigToM \cite{gandhi2023understandingsocialreasoninglanguage} have been introduced. \citet{sap2023neuraltheoryofmindlimitssocial} examines both false-belief understanding and social intelligence to determine if LLMs exhibit Theory of Mind abilities. They found that GPT-3’s social intelligence, as measured by SocialIQA, lags behind humans by over 30\%, and the model struggles with TOMI questions about mental states (55-60\%) compared to factual questions (90-100\%). \citet{liu2024interintentinvestigatingsocialintelligence} designs a benchmark to measure social intelligence based on LLMs' ability to understand intentions. They propose the Avalon game, an intention-guided social multi-agent game environment based on LLMs. Consistent with other studies, their results show that LLMs' ability to infer the intentions of others lags behind human performance by 20\%. \citet{cross2024hypotheticalmindsscaffoldingtheory} likewise evaluate LLMs' ability to infer other agents' strategies, goals, and capabilities. They similarly claim that while GPT-4 yields promising results, other models lack in performance, including GPT-3.

\paragraph{Human-Computer Interaction.} The work of \citet{Verma_2024} focus on human-robot interaction. Their observations indicate that even though LLMs exhibit behavior resembling ToM, these abilities are largely illusory, stemming from advanced language processing rather than a genuine understanding of mental states.

\paragraph{AI Psychology.} \citet{li2024quantifyingaipsychologypsychometrics} find that even though LLMs can handle basic ToM tasks, they perform inconsistently in more advanced scenarios requiring understanding of others' mental states. They notice that LLMs lack the flexibility needed to adapt to varied social contexts. Additionally, \citet{pi2024dissectingullmanvariationsscalpel} suggest that LLMs' information processing is more than mere pattern matching, but is still less sophisticated compared to human ToM abilities.

\paragraph{Negotiation.} \citet{chan2024negotiationtombenchmarkstresstestingmachine}, \citet{bianchi2024llmsnegotiatenegotiationarenaplatform} and \citet{kwon2024llmseffectivenegotiatorssystematic} investigate ToM abilities via negotiation. While \citet{chan2024negotiationtombenchmarkstresstestingmachine} focus on natural conversations and real-world data for evaluation, \citet{bianchi2024llmsnegotiatenegotiationarenaplatform} propose a game called NegotiationArena in which LLMs can compete against each other. \citet{chan2024negotiationtombenchmarkstresstestingmachine} conclude that current state-of-the-art LLMs perform significantly worse than humans, even when using advanced reasoning techniques. \citet{kwon2024llmseffectivenegotiatorssystematic} similarly report results below human baselines on partner modeling tasks in negotiation dialogues. \citet{bianchi2024llmsnegotiatenegotiationarenaplatform}, on the other hand, observe some tactics used by LLMs to improve their success but also note some irrational behavior.

\paragraph{Multi-agent Environment.} Several works on Machine Theory of Mind \cite{Li_2023, guo2023suspicionagentplayingimperfectinformation, zhou2023i, pmlr-v162-sclar22a, cross2024hypotheticalmindsscaffoldingtheory} have employed multi-agent collaboration or competition to assess the ToM skills of LLMs within symmetric environments, where multiple agents interact socially with one another and physically with their surroundings, thereby satisfying the requirements for situatedness. \citet{Li_2023} designs a game where 3 agents try to locate and safely defuse color-coded bombs scattered in an unexplored environment. They take action in communicating with each other and update their beliefs accordingly. \citet{guo2023suspicionagentplayingimperfectinformation} let the models play Leduc Hold’em, a two-player poker game. Models try to understand opponent's current beliefs and plan their actions accordingly. \citet{zhou2023i} investigate LLM interactions in the role-playing game Dungeons and Dragons (D\&D) using two agents: a teacher and a student. The teacher acts as the Dungeon Master, guiding the student to encourage specific actions that drive the game forward. The authors explore the LLMs' ability to generate this guidance as a proxy for Theory of Mind capabilities. \citet{pmlr-v162-sclar22a} presents SymmToM, which is a completely symmetric multi-agent environment in which all agents possess the ability to see, hear, speak, move, and actively participate. Successfully navigating SymmToM requires agents to demonstrate varying degrees of theory of mind and communicate effectively using a basic channel with a fixed set of symbols. Lastly, \citet{cross2024hypotheticalmindsscaffoldingtheory} evaluate LLMs' capability to compete and collaborate in several Multi-Agent Reinforcement Learning benchmarks, such as the Melting Pot evaluation suite \citep{leibo2021scalableevaluationmultiagentreinforcement}.

\paragraph{Multimodal ToM.} \citet{etesam2024contextualemotionrecognitionusing} test the ability of the models to recognize emotions from images. They show that LLMs with vision modules are more promising than other techniques, and the performance increases after fine-tuning on the data. \citet{chen2024theorymindseyereading} likewise claim that ToM tasks are temporal in essence, and thus videos are a suitable medium for evaluating ToM capabilities of LLMs. They report improved results with their novel fine-tuning method but conclude that the results are not robust, possibly due to the scarcity of related video datasets.

\paragraph{Meta Learning.}
Meta-learning in Theory of Mind with large language models refers to the models' ability to quickly understand and generalize well to new tasks with small data, akin to how humans make reasonable judgments based on limited information. \citet{yu2024fewshotcharacterunderstandingmovies} propose a new dataset called ToM-in-AMC, where models are expected to understand new characters from new movies based on characters they have seen in different films. The authors demonstrate that all their meta-learning approaches, including those based on GPT-4, lag behind human performance by 20\%. Similarly, \citet{rabinowitz2018machinetheorymind} frame the problem as a meta-learning task and demonstrate through a series of experiments that their Theory of Mind neural network learns a general model for agents within the training distribution, as well as how to build an agent-specific model online by observing a new agent’s behavior.

\paragraph{Prompting Frameworks.}
One research focus in Machine Theory of Mind involves developing various prompting strategies to improve model performance. Notable works in this area include \citet{wilf2023thinktwiceperspectivetakingimproves, hou-etal-2024-timetom, jung2024perceptionsbeliefsexploringprecursory, zhou2023farlargelanguagemodels, xu2024faithfullogicalreasoningsymbolic,wang2024metacognitivepromptingimprovesunderstanding, huang2024notioncomplexitytheorymind}.

\section{Discussion}
\label{discussion}

\subsection{Prompting for ToM in LLMs}
\label{prompt}

During LLM testing, various prompting strategies can enhance model performance. For instance, in \citet{brown2020languagemodelsfewshotlearners}, the authors demonstrate that larger models like GPT-3 significantly benefit from few-shot learning, where the model is provided with a few examples of the task during inference, enabling it to understand the desired output format better and improve its responses without additional fine-tuning. Motivated by this observation, prompting strategies are frequently employed in Machine Theory of Mind literature, including traditional methods such as few-shot learning \cite{brown2020languagemodelsfewshotlearners} and chain-of-thought prompting \cite{wei2023chainofthoughtpromptingelicitsreasoning}, as well as novel approaches specifically designed for Theory of Mind tasks, such as FaR (Foresee-and-Reflect) \cite{zhou2023farlargelanguagemodels} and TimeToM \cite{hou-etal-2024-timetom}.

In Table \ref{table:prompts}, we list the commonly used prompting techniques and novel methods proposed for Machine Theory of Mind, with brief explanations and references to example works using these methods. In general, few-shot prompting appears to enhance the performance of LLMs in Theory of Mind tasks, as demonstrated by \citet{wu-etal-2024-coke, kwon2024llmseffectivenegotiatorssystematic, moghaddam2023boostingtheoryofmindperformancelarge}. However, \citet{sap-etal-2022-neural} found that increasing the number of demonstrations does not continuously improve accuracy and tends to plateau at a certain point, likely due to recency bias—where the model eventually ``defaults to the most recent object location''. Chain-of-thought is another prompting technique where the model is guided to break down complex tasks into a sequence of intermediate reasoning steps. This approach helps the model generate more accurate and coherent responses, especially for tasks requiring multi-step reasoning. \citet{moghaddam2023boostingtheoryofmindperformancelarge} combines few-shot with CoT and clearly shows that various types of 2-shot CoT prompts significantly boost LLM performance in Theory of Mind tasks, compared to zero-shot prompting.

On the other hand, \citet{mireshghallah2024llmssecrettestingprivacy} highlights that using zero-shot CoT does not improve privacy, and in fact, it makes leakage more severe. Both GPT-4 and ChatGPT experience increased leakage when prompted with CoT. Moreover, \citet{chen2024tombenchbenchmarkingtheorymind} points out that task- or ability-oriented zero-shot CoT prompting cannot improve LLM performance (on 10 different models including GPT-4) or may even lead to worse results if the model lacks essential Theory of Mind abilities. Specifically, across ten LLMs, the average performance in English decreased by 2.1\% when CoT was employed. \citet{kim2023fantombenchmarkstresstestingmachine} reaches a similar conclusion, noting that CoT has a selective impact on performance, with improvements seen only in specific scenarios, such as ``reasoning specifically for determining characters who are unaware of certain information.'' Consistent with previous findings, \citet{gandhi2023understandingsocialreasoninglanguage} reports that zero-shot chain-of-thought (CoT) prompting doesn't consistently enhance performance across different conditions. However, introducing a one-shot CoT example results in reliable performance improvements across all conditions. \textbf{CoT with in-context examples boosts performance in ToM tasks, but zero-shot CoT often fails to improve results and can even lead to worse outcomes for LLMs lacking ToM abilities.}

Traditional prompting methods like Tree-of-Thoughts \cite{yao2023treethoughtsdeliberateproblem} and Self-Ask \cite{press2023measuringnarrowingcompositionalitygap} are less explored in Theory of Mind literature compared to few-shot and CoT techniques. \citet{zhou2023farlargelanguagemodels} evaluate CoT, ToT, Self-Ask, and their method, FaR (Foresee-and-Reflect), showing that FaR outperforms the others in most tasks. Similarly, \citet{huang2024notioncomplexitytheorymind} find that their proposed approach exceeds the performance of CoT and ToT alternatives.

Novel approaches are being explored as alternatives to traditional prompting techniques. \citet{wilf2023thinktwiceperspectivetakingimproves} introduce SimToM, which divides the task into two steps: first, the model determines what the person knows, and then it answers the question from that perspective. Generally, SimToM achieves better results on the ToMI \cite{le-etal-2019-revisiting} and BigTOM \cite{gandhi2023understandingsocialreasoninglanguage} datasets for both Llama and GPT-based models. Following this, \citet{hou-etal-2024-timetom} introduces TimeToM, a prompting technique that incorporates timelines into stories. They demonstrate that TimeToM outperforms zero-shot, zero-shot CoT, and SimToM on the same datasets using the same models. PercepToM \cite{jung2024perceptionsbeliefsexploringprecursory} is another method that prompts the model to infer characters' perspectives and answer the query based on the isolated context. \citet{zhou2023farlargelanguagemodels} introduces a novel zero-shot prompting framework, FaR (Foresee-and-Reflect), demonstrating its superiority over CoT and Self-Ask on the new T4D (Thinking for Doing) task formulation. FaR increases GPT-4's accuracy from a baseline of 50\% to 71\% and improves other LLMs by 12\% to 18\%. The authors also evaluate their models on Faux-Pas and Story Structure Tests to assess the generalizability of their approach.

To initiate the reasoning process, \citet{xu2024faithfullogicalreasoningsymbolic} propose a logic-based approach that translates natural language into a symbolic format, enabling reasoning through formal logical rules. Their results demonstrate that SymbCoT significantly enhances the reasoning capabilities of the vanilla CoT.

Inspired by studies on human introspective reasoning, \citet{wang2024metacognitivepromptingimprovesunderstanding} introduce Metacognitive Prompting (MP). MP approaches problem-solving through five consecutive steps: comprehension clarification, preliminary judgment, critical evaluation, decision confirmation, and confidence assessment. This method outperforms Chain of Thought (CoT) across both 0-shot and 5-shot settings on ten natural language understanding (NLU) datasets from the GLUE, SuperGLUE, BLUE, and LexGLUE benchmarks. Another approach with the same name, proposed by \citet{leer2023violationexpectationmetacognitiveprompting}, prompts the model to generate 'thoughts' about a given task and then uses those 'thoughts' as the context in subsequent inference steps.

Lastly, the DWM (Decomposed World Model), proposed by \citet{huang2024notioncomplexitytheorymind}, involves interactively guiding a language model through a Theory of Mind problem. The process starts by asking the model to generate concise representations of each agent's beliefs. The input is then divided into $T$ state descriptions, and the model is prompted to describe how each state and the environment evolve. Finally, all input segments and descriptions are combined and re-fed into the model with an additional prompt to produce the final task solution. The authors compare the performance of DWM to CoT across five different Theory of Mind benchmarks: ToMi \cite{le-etal-2019-revisiting}, FANToM \cite{kim2023fantombenchmarkstresstestingmachine}, Mindgames \cite{sileo-lernould-2023-mindgames}, Adv-CSFB \cite{shapira2023cleverhansneuraltheory}, and SocialIQA \cite{sap-etal-2019-social}. DWM outperforms CoT on FANToM, Mindgames, and Adv-CSFB but falls short on ToMi and SocialIQA.

Nearly all the novel prompting strategies designed to enhance the Theory of Mind performance of language models have emerged in the past two years, highlighting a competitive race in the literature. \textbf{While these strategies have demonstrated superior performance compared to CoT on several datasets, there remains a notable gap in the literature regarding their comparative effectiveness against one another.}

\newcolumntype{P}[1]{>{\centering\arraybackslash}p{#1}}
\begin{table*}[h]
\small
\centering
  
  \label{table:intrinsic_dataset}
  \begin{tabular}{P{3cm}|P{7.5cm}|P{4.3cm}}
   \toprule
    \textbf{Technique}&\textbf{Brief Description}&\textbf{Examples of Machine ToM Applications}\\
    \midrule
Few-shot prompting \cite{brown2020languagemodelsfewshotlearners} & 
It involves providing a model with only a few (input, output) pairs as demonstrations in the prompt. This allows the model to generalize from minimal examples and perform the task with limited data. & \cite{yu2024fewshotcharacterunderstandingmovies, sap2023neuraltheoryofmindlimitssocial, gandhi2023understandingsocialreasoninglanguage, kwon2024llmseffectivenegotiatorssystematic, cross2024hypotheticalmindsscaffoldingtheory, moghaddam2023boostingtheoryofmindperformancelarge}, \\
\midrule
     Chain-of-thought (CoT) prompting \cite{wei2023chainofthoughtpromptingelicitsreasoning}  & Prompting language model to generate 
a series of intermediate natural language reasoning steps that lead to the final output. & \cite{ shapira2023cleverhansneuraltheory, kim2023fantombenchmarkstresstestingmachine, chen2024tombenchbenchmarkingtheorymind, chan2024negotiationtombenchmarkstresstestingmachine, sabour-etal-2024-emobench} \\
     \midrule
     Tree-of-thought (ToT) prompting \cite{yao2023treethoughtsdeliberateproblem}  & It is a reasoning approach that explores multiple potential solution paths, branching out like a tree, to evaluate different outcomes and make decisions. & \cite{huang2024notioncomplexitytheorymind, zhou2023farlargelanguagemodels}\\
     \midrule 
     Self-Ask prompting \cite{press2023measuringnarrowingcompositionalitygap} &  The model explicitly asks and answers follow-up questions before addressing the initial question. & \cite{zhou2023farlargelanguagemodels}\\
     \midrule
     SimToM \cite{wilf2023thinktwiceperspectivetakingimproves} & The LLM filters the context to only what the character knows, then answers the question based on this limited information. & \cite{wilf2023thinktwiceperspectivetakingimproves, hou2024timetomtemporalspacekey} \\
     \midrule
     TimeToM \cite{hou2024timetomtemporalspacekey} & It creates a timeline for stories or dialogues and builds a Temporal Belief State Chain (TBSC) for each character based on the events they know. & \cite{hou2024timetomtemporalspacekey} \\
     \midrule PercepToM \cite{jung2024perceptionsbeliefsexploringprecursory} & It infers which characters perceive each piece of information and extracts the perspective context for the target character by string-matching the inferred perceptions. Then the LLM generates a response based on this filtered context. & \cite{jung2024perceptionsbeliefsexploringprecursory}\\
     \midrule FaR (Foresee and Reflect) \cite{zhou2023farlargelanguagemodels} & This framework enables LLMs to predict future events from observations and determine actionable steps to assist humans in real-time contexts. & \cite{zhou2023farlargelanguagemodels} \\
     \midrule
     SymbCoT (Symbolic Chain-of-Thought) \cite{xu2024faithfullogicalreasoningsymbolic} & It translates natural language context into a symbolic format, creates a step-by-step plan using symbolic logic, and then verifies the translation and reasoning. & \cite{wang2024symbolicworkingmemoryenhances, xu2024faithfullogicalreasoningsymbolic} \\
     \midrule
     Metacognitive prompting\footnotemark  
 \cite{wang2024metacognitivepromptingimprovesunderstanding} & It addresses the query in five key steps: (1) Clarify the understanding of the question, (2) Provide a preliminary interpretation, (3) Critically assess the initial interpretation, (4) Finalize the decision and provide reasoning, and (5) Evaluate your confidence. & \cite{wang2024metacognitivepromptingimprovesunderstanding, Ohtani2024DoesMP} \\
     \midrule
     Discrete World Models (DWM) prompting \cite{huang2024notioncomplexitytheorymind} & It divides the input into $T$ state descriptions. For each, the LLM provides the state event and its changes. Finally, it combines the input and descriptions to prompt the LLM for the task answer. & \cite{huang2024notioncomplexitytheorymind} \\

\bottomrule
\end{tabular}
\captionsetup{font=normalsize}
\caption{\label{table:prompts}Prompting frameworks used in Machine Theory of Mind.}
\end{table*}

\footnotetext{There is also a less well-known method introduced by \citet{leer2023violationexpectationmetacognitiveprompting} with the same name.}

\subsection{Fine-tuning LLMs for ToM}
\label{finetuning}

Fine-tuning models for various downstream tasks is an alternative to looking for an optimal prompt. It is possible to adapt LLMs to a specific task using a smaller training set \cite{howard2018universallanguagemodelfinetuning}. Reinforcement learning is another tool that has been used for developing the most advanced models \cite{ouyang2022traininglanguagemodelsfollow, bai2022traininghelpfulharmlessassistant, pmlr-v162-sclar22a}. Reinforcement learning allows us to train LLMs that are aligned with the human preferences. Both supervised fine-tuning and reinforcement learning have been used for improving the ToM abilities of the models.

Supervised fine-tuning has been the popular choice for ToM. 
\citet{wu-etal-2024-coke}, 
\citet{kim2023fantombenchmarkstresstestingmachine} and \citet{tang2024tomlmdelegatingtheorymind} use a small supervised set for improving ToM abilities on natural language generation. \citet{cohen2021exploring} fine-tunes RoBERTa for natural language inference and reports ToM performance. Additionally, \citet{etesam2024contextualemotionrecognitionusing} and \citet{chen2024theorymindseyereading} perform supervised fine-tuning on multimodal tasks. Finally, \citet{zhou2023i} show an example of reinforcement learning fine-tuning for ToM. They aim to guide LLMs by modeling intents and theory of mind.

An important concern is the robustness of the fine-tuned models on ToM tasks. The evaluation examples are mostly very similar to the training data, and the reported results can therefore be misleading. Testing models with different data is necessary to have a correct estimation of their abilities.

\subsection{LLMs’ systematic failures}
\label{failure}

One valuable practice is to conduct error analysis on misclassified examples to identify systematic failures in large language models. This approach enables future research to target and address specific issues more effectively.

One type of error, highlighted by \citet{ma2023tomchallengesprincipleguideddatasetdiverse}, is \textbf{conservative errors}, where the model fails to infer an agent's belief, stating that there isn't enough information. An example response to this is given as follows: \enquote{\textit{The context does not provide information on where
Juanita would look for the towel when she returns.}} At times, the model may refuse to answer by stating that none of the provided options are correct \cite{sabour-etal-2024-emobench}. \citet{ma2023tomchallengesprincipleguideddatasetdiverse} also discuss \textbf{hallucination errors}, where the model provides answers that include information not inferred from the narrative but instead fabricated by the model. Hallucination is a significant issue not only in Machine Theory of Mind \cite{Li_2023, liu2024interintentinvestigatingsocialintelligence, ma2023tomchallengesprincipleguideddatasetdiverse, guo2023suspicionagentplayingimperfectinformation, bianchi2024llmsnegotiatenegotiationarenaplatform, wu-etal-2023-hi} but across all areas of the LLM literature \cite{huang2023surveyhallucinationlargelanguage}. \citet{xu2024hallucinationinevitableinnatelimitation} claims that the hallucination is inevitable for LLMs, pointing out several attempts to mitigate the effect such as factual-nucleus sampling \cite{lee2023factualityenhancedlanguagemodels}, chain-of-verification \cite{dhuliawala2023chainofverificationreduceshallucinationlarge}, retrieval-augmentations \cite{lewis2021retrievalaugmentedgenerationknowledgeintensivenlp}. \citet{Li_2023} also addresses the issue of hallucination errors in their study. They attribute these errors primarily to the absence of explicit belief representation. To explore this issue, they evaluate the GPT-4+Belief condition, where the models record their belief states in text. Their findings reveal that incorporating explicit belief states reduces invalid actions by 50.7\% and boosts team efficiency by 130\% in a multi-agent collaboration environment.

Another problem mentioned by \citet{Li_2023} is the \textbf{long-horizon context}, where a model tends to forget information about the room and some details in the inquiry text as they got far away from the planning question at the very end. 

\citet{wu-etal-2023-hi} lists five error types observed in their experiments: insufficient reasoning depth, commonsense errors, hallucinations, temporal ignorance, and spurious causal inference.

\textbf{Insufficient reasoning depth} refers to an LLM skipping essential reasoning steps to answer a lower-order question (e.g., responding to 'Where is Anne?' instead of 'Where would John think Anne is?'). \textbf{Commonsense errors} occur when the model generates a continuation that defies common sense. For instance, the authors provide an example where someone leaves a closed space yet can still see what is happening inside. \textbf{Temporal ignorance} occurs when LLM ignores or confuses the order of the events. \textbf{Spurious causal inference} refers to incorrectly attributing a cause-and-effect relationship between events that are not actually related in that way.

\subsection{Major Challenges in Evaluation}
\label{challenges}

Evaluating large language models on Theory of Mind tasks presents significant challenges for researchers. Recent benchmarks and ToM papers have highlighted several of these issues.

\paragraph{Training Data Bias.} One significant issue, not limited to ToM evaluations, is the inherent bias in the training data, which can affect the fairness and accuracy of large language models. These biases can manifest across various domains, including age, profession, nationality, race, religion, gender, socioeconomic status, and political ideology. Several datasets are introduced to measure the bias in language models such as Bias in Open-Ended Language
Generation Dataset (BOLD) \cite{Dhamala_2021}, the Bias Benchmark for QA \cite{parrish2022bbqhandbuiltbiasbenchmark} and the Holistic Bias dataset \cite{smith-etal-2022-im}. Recently, additional categories of bias have been identified, such as demographic biases \cite{ma-etal-2023-deciphering} and geographical biases \cite{manvi2024largelanguagemodelsgeographically}.

To mitigate the effects of bias, several strategies have been proposed, including counterfactual data augmentation \cite{zmigrod-etal-2019-counterfactual}, where for example, new training sentences are generated by replacing gendered words with their opposites; word embedding arithmetic \cite{bolukbasi2016mancomputerprogrammerwoman}, which removes the vector component corresponding to the bias source from the original vector; dropout regularization \cite{webster2021measuringreducinggenderedcorrelations}; Iterative Null-space Projection \cite{ravfogel-etal-2020-null}, involving training a linear classifier to predict the bias term and then projecting representations onto the classifier's null space; self-debiasing \cite{schick2021selfdiagnosisselfdebiasingproposalreducing}, which leverages LLMs to detect and reduce bias; Sent-Debias \cite{liang-etal-2020-towards}, which computes and removes the bias subspace from the original space; down-weighting the biased data \cite{karimi-mahabadi-etal-2020-end, utama-etal-2020-towards}, debiasing using contrastive learning \cite{lyu2023featureleveldebiasednaturallanguage} and domain adaptation \cite{ranaldi2023tripfairnessbiasdebiasing}, where LLMs are fine-tuned—potentially through methods like LoRA \cite{hu2021loralowrankadaptationlarge}—to reduce bias within specific domains.

\paragraph{Prompt Bias.} Prompt engineering is a crucial technique for enhancing the performance of LLMs without modifying their pre-trained parameters. The performances of language models in Theory of Mind tasks have been demonstrated to improve through various prompting techniques, such as few-shot learning \cite{sap-etal-2022-neural} and chain-of-thought prompting \cite{moghaddam2023boostingtheoryofmindperformancelarge}.

However, prompts can introduce bias, complicating the fair evaluation of LLMs. For instance, in few-shot learning, the order of examples \cite{lu2022fantasticallyorderedpromptsthem, robinson2023leveraginglargelanguagemodels} can influence a model toward a specific prediction. One type of this bias is position bias \cite{zheng2023judgingllmasajudgemtbenchchatbot}, where the input sequence affects the model's decision. A specific form of position bias is recency bias, where the model tends to repeat answers that appear at the end of the prompt \cite{zhao2021calibrateuseimprovingfewshot}.

The choice of tokens (A-D) for different options can also introduce bias if the model has a prior preference for one token over another (common token bias) \cite{zheng2024largelanguagemodelsrobust, zhao2021calibrateuseimprovingfewshot}. \cite{chen2024tombenchbenchmarkingtheorymind} addresses this issue by shuffling the order of options five times and selecting the most frequently chosen option as the final answer. Additionally, the format of the prompt and the selection of examples for in-context learning can impact evaluation outcomes \cite{zhao2021calibrateuseimprovingfewshot}. Another issue is majority label bias, where a model tends to favor more frequent answers in the prompt \cite{zhao2021calibrateuseimprovingfewshot}. When evaluating LLMs with LLMs, other issues may arise, such as verbosity bias \cite{Saito2023VerbosityBI,zheng2023judgingllmasajudgemtbenchchatbot}, where an LLM judge is more likely to prefer a verbose output even if a shorter one is better, or self-enhancement bias, where an LLM judge is more likely to favor output it generated itself.
Finally, when exemplars have equally predictive labels, a model might exhibit bias toward a particular feature, such as sentiment over lexical features, as shown by \cite{si2023measuringinductivebiasesincontext}. Challenges that arise when using LLMs as judges are particularly significant for Theory of Mind evaluations, as language models may be employed for auto-grading \cite{ma2023tomchallengesprincipleguideddatasetdiverse} for open-ended tasks.

One approach to improve the few-shot performance by mitigating prompt bias is called contextual calibration \cite{zhao2021calibrateuseimprovingfewshot}, where we try to learn an affine transformation which maps the predicted outputs to a uniform distribution if applied on the \textit{context-free} input. A recent alternative approach to affine transformation is removing the biased component (identified through prompt-only querying)
from the representation vectors \cite{xu2024carepromptbiasinvestigating}. Some of the techniques originally designed to mitigate pre-training data bias can also be applied to reduce prompt bias in language models.

\paragraph{Training contamination.} Another significant challenge is training contamination \cite{hagendorff2024machinepsychology}, which occurs when models are exposed to tasks and prompts from existing literature during training, potentially skewing results. This issue is particularly problematic in the psychological domain, where researchers often use prompts directly from existing studies, leading to biased outcomes.
\cite{ullman2023largelanguagemodelsfail} even suggests that introducing a systematic task generator might backfire on future researchers, as it allows LLMs to prepare for evaluations by memorizing large volumes of examples.

Several efforts have been made to identify pre-training data to determine whether a model has encountered specific queries during training. One approach measures the changes in log probabilities when the order of input text is altered \cite{oren2023provingtestsetcontamination}. Heuristically, a greater change in log probability suggests that the model favors a specific order it may have seen during training. Another approach involves measuring the log-likelihood of the input text and applying a threshold to identify if the text is ''too likely``, indicating that it may have appeared in the corpus \cite{shi2024detectingpretrainingdatalarge}. Additionally, researchers urge the LLM developers to provide necessary tools along with the models to test the membership of any input in the pre-training data \cite{marone2023dataportraitsrecordingfoundation}.

To tackle the contamination problem, researchers and practitioners are recommended to curate their evaluation datasets or use procedurally generated datasets like CogBench \cite{codaforno2024cogbenchlargelanguagemodel} where the task content is generated often on the fly using predefined rules. Detecting pre-training data contamination is a vibrant research area that could help the community mitigate the effects of training data contamination.

\paragraph{Limited scope of ToM evaluations}
Over-generalized conclusions pose a significant challenge in the evaluation of LLMs \cite{ma-etal-2023-towards-holistic, shapira2023cleverhansneuraltheory}, as researchers may focus on a specific category (or subcategory) to assert that ToM has emerged in LLMs. 
\citet{shapira2023cleverhansneuraltheory, ullman2023largelanguagemodelsfail} both demonstrate that small variations in the original queries can lead to significant changes in model performance, indicating that these models do not possess \textit{robust} ToM abilities.

\paragraph{Manual scoring} In addition to tasks with constrained formats (e.g., true or false questions), many existing Theory of Mind tasks involve open-ended questions that require manual scoring by the experimenter \cite{Beaudoin2020}. This method is often impractical for evaluating large language models, as it demands significant human effort \cite{chen2024tombenchbenchmarkingtheorymind} and introduces potential biases in judgment \cite{klie2022annotationerrordetectionanalyzing}. Therefore, an auto-grader is essential to automate the evaluation process for large volumes of tasks \cite{ma2023tomchallengesprincipleguideddatasetdiverse}.

\paragraph{Shortcut learning.} A spurious correlation refers to the co-occurrence of certain cues, features, or labels that are not causally related to the task. Shortcut learning \cite{Geirhos_2020} occurs when a model relies on such non-essential cues, which are spuriously correlated with the task, rather than focusing on the actual, relevant features needed for accurate predictions. Evaluating Theory of Mind in LLMs is also prone to spurious correlations, where models may learn shortcuts instead of genuine ToM skills \cite{Aru_2023, shapira2023cleverhansneuraltheory}. These shortcuts can create the illusion that language models possess ToM abilities, even when they do not \cite{dziri2023faithfatelimitstransformers}.

Additionally, challenges unique to Theory of Mind evaluations include the inherent complexity of mental state reasoning, which requires models to navigate nuanced, context-dependent human emotions and intentions. Current benchmarks may also be limited in their ability to fully capture the depth of Theory of Mind understanding in AI systems. A summarized version of the major challenges and proposed solutions can be found in Table \ref{tab:challenge}.

\section{Conclusion}
\label{conclusion}

In this work, we gather and summarize the latest advancements and discussion on the ToM abilities of LLMs. We list the benchmarks and results analyzing different models, providing a comprehensive overview of their current capabilities. Our findings suggest the following narrative: As language models become more advanced—incorporating more parameters and larger training datasets—they tend to achieve higher scores on Theory of Mind tasks compared to their predecessors \cite{van-duijn-etal-2023-theory, shapira2023cleverhansneuraltheory, hou-etal-2024-timetom, kosinski2024evaluatinglargelanguagemodels}. However, in most cases, they still fall short of human performance \cite{mireshghallah2024llmssecrettestingprivacy, liu2024interintentinvestigatingsocialintelligence, zhou2023farlargelanguagemodels, sap2023neuraltheoryofmindlimitssocial}. Consequently, we align with the skeptical perspective, suggesting that while LLMs may exhibit some enhanced ToM abilities, these capabilities remain limited and may often rely on spurious correlations rather than robust understanding. With this review of the current state of Machine Theory of Mind literature, we aim to pave the way for future research focused on enhancing the ToM capabilities of LLMs and expanding their applications across diverse domains.

\subsection{Appendices}

\begin{table*}[!hbtp]
    \centering
    \small
    \begin{tabular}{cp{1in}p{2.3in}p{2.6in}}
    \toprule
    \multicolumn{2}{c}{Challenges in Evaluation}  & \multicolumn{1}{c}{Description} & \multicolumn{1}{c}{Example Solution Approaches} \\\midrule
    {$\blacktriangleright$} & \emph{Training Data Bias}
    &  Inherent biases in the model's pre-training data span various domains, including age, race, religion, and socioeconomic status.
    &  Counterfactual data augmentation \cite{zmigrod-etal-2019-counterfactual}, word embedding arithmetic \cite{bolukbasi2016mancomputerprogrammerwoman}, dropout regularization \cite{webster2021measuringreducinggenderedcorrelations}, null-space projection \cite{ravfogel-etal-2020-null} \\\midrule
    
    {$\blacktriangleright$} & \emph{Prompt Bias}
    & Unintended influence or skewing of a language model's outputs based on how a prompt or question is phrased can manifest in various forms, such as majority label bias, position bias, recency bias, common token bias, and contextual cultural or societal biases. & Contextual calibration \cite{zhao2021calibrateuseimprovingfewshot}, debiasing via representation vector manipulation \cite{xu2024carepromptbiasinvestigating} or basic approaches like shuffling the options couple of times \cite{chen2024tombenchbenchmarkingtheorymind}.
    \\\midrule
    
    {$\blacktriangleright$} & \emph{Training\newline contamination}
    & Models are already exposed to tasks and prompts from existing literature
during training, potentially skewing results
    & Generating new datasets for experiments, using procedurally generated benchmarks \cite{codaforno2024cogbenchlargelanguagemodel}, detecting pretraining data to find out possible exposure \cite{shi2024detectingpretrainingdatalarge}
    \\\midrule
    
    {$\blacktriangleright$} & \emph{Shortcut learning\newline(Spurious\newline correlations)}
    &  The phenomenon where a model learns to solve a task by exploiting simple, superficial patterns or correlations in the data, rather than learning the underlying concepts or more robust features needed for generalization. This presents a challenge for evaluation as learning the shortcuts can maximize the performance metrics without really understanding the task.
    &  Conducting experiments on more and diverse datasets \cite{zhou2023farlargelanguagemodels}, cross-bias generalization techniques such as Group Distributionally Robust Optimization \cite{sagawa2020distributionallyrobustneuralnetworks}, Domain-Adversarial Training of Neural Networks (DANN) \cite{ganin2016domainadversarialtrainingneuralnetworks},
    Learning from Failure \cite{nam2020learningfailuretrainingdebiased}, ReBias: Representational Regularization \cite{bahng2020learningdebiasedrepresentationsbiased}
    \\\midrule

    {$\blacktriangleright$} & \emph{Subjectivity in\newline Scoring}
    & Manual scoring can lead to subjective evaluations, as different assessors may interpret responses differently based on their own biases, experiences, or expectations. This can result in inconsistencies and variability in scores, making it challenging to objectively compare results across different studies or models. 
    &  Preparation of benchmarks and task formulations suited for automatic grading \cite{ma2023tomchallengesprincipleguideddatasetdiverse}, Increasing the number of annotators or graders to achieve robust evaluations and presenting standardized criteria. 
    \\\midrule 
    
    {$\blacktriangleright$} & \emph{Limited scope of ToM evaluations}
    & Theory of Mind encompasses a wide range of mental states, as categorized by \citet{Beaudoin2020}. However, much of the existing literature focuses on specific tasks, making it difficult to generalize the findings.
    & Integrating more ATOM mental states into benchmarks and task formulations, along with considering different situatedness, as suggested by \citet{cao-2021-holistic}. Experimenting on different benchmarks \cite{huang2024notioncomplexitytheorymind, shapira2023cleverhansneuraltheory, hou-etal-2024-timetom},  \\
    
    \bottomrule
    \end{tabular}
    \caption{Types of evaluation challenges commonly discussed in the literature, along with their descriptions and example solution approaches.}
    \label{tab:challenge}
\end{table*}

\clearpage
\bibliography{custom}
\bibliographystyle{acl_natbib}

\end{document}